\documentclass[11pt,oneside,a4paper]{article}
\usepackage{graphicx}
\usepackage{setspace}
\usepackage{lineno}
\usepackage{algorithm}
\usepackage{algpseudocode}
\usepackage{amsmath}
\usepackage[justification=centering]{caption}
\usepackage{color,soul} 
\title{Visual Perception and Modelling in Unstructured Orchard for Apple Harvesting Robots}
\date{}
\author{Hanwen Kang and Chao Chen$^{*}$}
\begin{document} 
	\maketitle
	%\linenumbers
	
	\begin{abstract}
		Vision perception and modelling are the essential tasks of robotic harvesting in the unstructured orchard. This paper develops a framework of visual perception and modelling for robotic harvesting of fruits in the orchard environments. The developed framework includes visual perception, scenarios mapping, and fruit modelling. The Visual perception module utilises a deep-learning model to perform multi-purpose visual perception task within the working scenarios;  The scenarios mapping module applies OctoMap to represent the multiple classes of objects or elements within the environment;  The fruit modelling module estimates the geometry property of objects and estimates the proper access pose of each fruit. The developed framework is implemented and evaluated in the apple orchards. The experiment results show that visual perception and modelling algorithm can accurately detect and localise the fruits, and modelling working scenarios in real orchard environments. The $F_{1}$ score and mean intersection of union of visual perception module on fruit detection and segmentation are 0.833 and 0.852, respectively. The accuracy of the fruit modelling in terms of centre localisation and pose estimation are 0.955 and 0.923, respectively. Overall, an accurate visual perception and modelling algorithm are presented in this paper.
	\end{abstract}
    

	\section{Introduction}
	Robotic harvesting technique shows a promising aspect in the future development of the agriculture industry. Machine vision is one of the essential element in the harvesting robots. The vision system of a harvesting robot is required to precept environment information, such as the location of target fruits and the surrounding environment, to guide the manipulator to perform the harvesting task. There are many factors which can heavily affect the performance of the vision system, including illumination changing, objects occlusion, and noisy background. Compared to perform robotic harvesting in the structured orchard which has been optimise-designed for automation purpose, robotic harvesting in unstructured orchard environments is even challenging. Since the complex arrangement of branches, leaves, and other objects can easily obstacle the detachment path of the manipulator, leading to the failure of operation or even damage the manipulator. Therefore, a robust and efficient perception and modelling algorithm is the crucial element to develop universal harvesting robots.
	
	In this work, a machine vision system for apple harvesting robots is presented. The developed vision system includes a multi-purpose Deep Network Detection And Segmentation Network (DASNet) for visual perception in 2D images and an environment modelling algorithm for information processing on 3D point clouds. The following highlights are presented in this paper:
	\begin{itemize}
		\item Development of a multi-purpose deep network DASNet, which combines the detection and instance/semantic segmentation in a one-stage detection network architecture, to perform the real-time environment perception in unstructured orchard environments.
		\item Development of an universal framework to perform visual perception and modelling in the unstructured orchard environments to guide robotic harvesting of fruits. 
		\item The developed perception and modelling vision algorithm were implemented and evaluated on the real unstructured orchard environment, providing a guideline for the designing of the robotic system in similar working environments.
	\end{itemize}
	
	The rest of the paper is organised as follow. Section 2 reviews the related works. Sections 3 and 4 introduce the methodology and experiment of the work, respectively. In section 5, the conclusions are presented.
	
	\section{Related Works}
	\subsection*{A: Visual Perception}
	Robotic vision in agriculture applications can apply different visual sensors, including RGB/stereo camera, RGB-D camera, Light Detection And Ranging (LiDAR), thermal imaging sensor, and spectral camera \cite{2016TMW_review2}. This work focus on reviewing the vision processing method on 2D RGB images. Vision perception of RGB images has been extensively studied. Traditional image processing algorithm applies feature descriptors to describe objects of interest within images, and then the machine-learning based algorithms are applied to perform the classification or detection accordingly. Many feature descriptors such as Colour Coherence Vector (CCV), Histogram of Gradient (HoG) and Scale Invariance Feature Transform (SIFT) were applied in the previous works. Correspondingly, traditional machine-learning based algorithms such as K-mean, Support Vector Machine (SVM), and Neural Network (NN) were applied to perform the classification on such descriptors. Nguyen et al. \cite{2016colorgeof} applied colour information and 3D geometry descriptor to perform detection of apple fruit with RGB-D camera. Lin et al. \cite{lin2019colorgeof} applied HSV colour feature and 3D geometry features on point clouds to describe the appearance of multiple classes of fruits, then an SVM classifier was trained to perform the detection based on such features. Wang and Xu \cite{wang2018unsupervised} applied multiple image feature descriptors and Latent Dirichlet Allocation (LDA) model to perform unsupervised segmentation of plants and fruits in greenhouse environments. Similar works of traditional machine vision algorithm in agriculture applications can also be referred to the reviews \cite{2012TMW_review1,2012TMW_review3}.
	
	More recently, deep-learning based algorithm shows advance and robust performance in the many tasks of machine vision. Deep-learning based algorithm applies deep Convolution Neural Network (CNN) architecture to perform automatic selection and learning of proper features from images. Many classic deep-learning architectures for different purposes were proposed. Region Convolution Neural Network (RCNN) \cite{2015faster-rcnn} and YOLO \cite{2018yolov3} are state of the art in object detection. The former method utilises two-stage detection strategies, applying a Region Proposal Network (RPN) to search Region of Interest (ROI) and a classification network to classify the objects and optimise the boundary box within ROI. YOLO applies fully convolution network architecture, combining the ROI searching and classification into a single step, reduce the computation complexity of the objects detection compared to the RCNN. Other works of the Deep CNN such as Fully Convolution Network (FCN) and Mask-RCNN for semantic segmentation and instance segmentation can be referred to the works \cite{2015FCN} and \cite{2017mask-rcnn}, respectively. In the machine vision in agriculture applications, the works of \cite{2016DLW_1} and \cite{2017DLW_2} applied faster-RCNN to perform detection of multiple classes of fruits in farm conditions, including apples, mango, and pepper. The works of \cite{2019DLW_3} and \cite{2019DLW_4} utilised the Mask-RCNN model to perform detection and instance segmentation in the application of robotic harvesting of strawberry fruits. The works of \cite{2019DLW_5} and \cite{2019DLW_6} applied YOLO on real-time in-field detection of apple and mango fruits for yield estimation and monitoring, respectively. The works of \cite{2019DLW_7} and \cite{2017DLW_8} applied FCN model to detect the guava fruits and cotton, respectively. In our previous works \cite{2019LedNet,2019DasNet-v1,2019DasNet-v2}, a YOLO and SPRNet \cite{2019SPRNet} architecture based multi-purpose deep CNN network model DASNet was developed to perform real-time detection, instance segmentation of fruits, and semantic segmentation of branch/trunk in orchard environments. More similar works of deep-learning based algorithms in machine vision in agriculture applications can also be found in the recent survey \cite{2018DLW_review}.
	 
	\subsection*{B: 3D Visual Mapping}
	3D mapping of the working scenarios is an essential element in many robotic applications. There are several implementations of 3D mapping methods which have been applied in the previous works. Tabak et al. \cite{1989mapping_1} discretised the mapping area of the 3D environment with equal size voxels. Voxelisation of 3D space requires large memory consumption when large outdoor scenarios and fine resolution are presented. With the development of the rang sensor, point cloud becomes another popular approach of mapping. Cole and Newman \cite{2006mapping_2} used 3D point clouds to present the 3D space in outdoor 3D SLAM system. However, the computation and memory consumption of this representation becomes enormous with increasing measurement resolution. For example, the depth camera with a resolution of 640 $\times$ 480 can output up to 300 thousand points. Other methods like elevation map \cite{2009mapping_3,2010mapping_4} or surface representations \cite{2007mapping_5,2007mapping_6} can only meet requirements when certain assumptions are made. Hornung et al. \cite{2013octomap} presented an octree-based 3D mapping approach OctoMap to represent the 3D space into a memory efficiency volumetric occupancy map. Also, such a method provides other advantages as adjustable resolution and high update frequency \cite{2018octomap_flight}. Therefore, OctoMap provides an efficient and robust approach to modelling the unstructured scenarios \cite{2018semanticOCTO}.
	
	There are several previous studies which have explored the environment modelling in the orchard scenarios. Adhikari and Manoj \cite{2011orchard_modelling_1} applied a Time-of-Flight (ToF) depth camera to perform visual sensing, and a 3D skeletonization algorithm was used to modelling the 3D profile of the branch/trunk for mechanical pruning. Wang and Zhang \cite{2013orchard_modelling_2} applied two RGB-D cameras to reconstruct dormant tree in the form of 3D pint cloud under laboratory environment. Amatya et al. \cite{2016orchard_modelling_3} applied a Bayesian-based classifier to perform classification of branch pixels within orchard environments, and then the branch was represented in the form of the straight line in the 2D images. Li et al. reconstructed the tree from the front view and back view, and the branch was modelled as a cylinder through the random sample consensus (RANSAC) algorithm. Lin et al. \cite{2019DLW_7} applied FCN to perform segmentation on branch and modelling it as the 3D line in the scenario to perform pose estimation of guava fruits. From our reviews, previous works of orchard modelling have several shortages in terms of robustness in real orchard environment, accuracy modelling, and inefficient computation. Therefore, a universal framework which includes OctoMap based modelling algorithm and corresponds fruit modelling algorithm is developed in this paper to perform visual perception and modelling in the unstructured orchard scenarios.  
	
	\section{Methodologies and Materials} 
	\subsection{System Architecture}
	\begin{figure}[h]
		\centering
		\includegraphics[width=\textwidth]{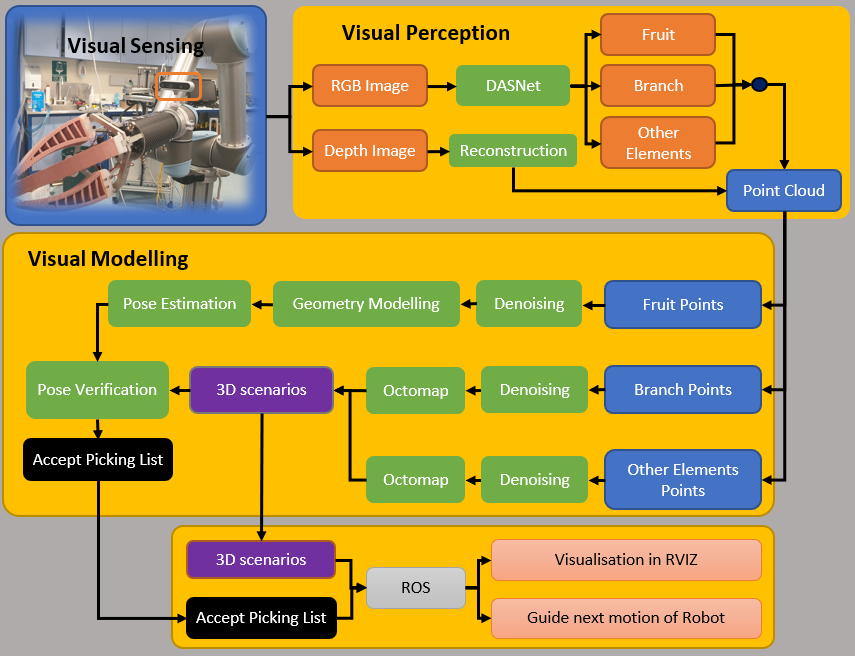}
		\caption{Overall architecture of visual perception and modelling subsystem of the designed harvesting robot.}
		\label{fig:work_flow}
	\end{figure}
	The developed apple harvesting robot contains a robot arm $\&$ gripping subsystem and a visual perception system. RGB-D camera (for example, Kinetic-v2 or RealSense D-435) are used to sensing the colour and depth information from the working scenarios. The applied RGB-D camera can be either installed on the gripper or at the back frame of the robot. In each running iteration of robotic control computation, visual perception and modelling algorithm is designed for sensing and process the visual information to guide the next action of the manipulator or gripper.
	
	The workflow of the visual perception and modelling algorithm is shown in Figure \ref{fig:work_flow}. Firstly, DASNet \cite{2019DasNet-v1,2019DasNet-v2} performs segmentation and detection on input RGB images to extract objects of interest from the working scenarios. Then, by combining the depth map which is collected by RGB-D camera, the processed information is used to modelling the working scenarios of the orchards. The branch/trunk or other elements (other possible obstacles) are represented in the form of the OctoMap. The fruits are represented in the form of the sphere, of which a 3D Sphere Hough Transform (3D-SHT) algorithm is used to estimate the geometry property of each fruit. Based on the point distribution, the accessible pose is estimated for each of fruits. Furthermore, to ensure the robustness of the pose estimation, a 3DVHF+ \cite{2014_3dvfh+} based pose verification algorithm is developed to check the confidence level of the estimated pose. Lastly, the acceptable picking list of fruits and 3D scenario mapping are sent to the centre control, which can decide and generate the next motion and action of manipulator or gripper.
	
	\subsection{Visual Perception of Environment}
	\subsubsection{Multi-purpose Network}
	\begin{figure}[h]
		\centering
		\includegraphics[width=\textwidth]{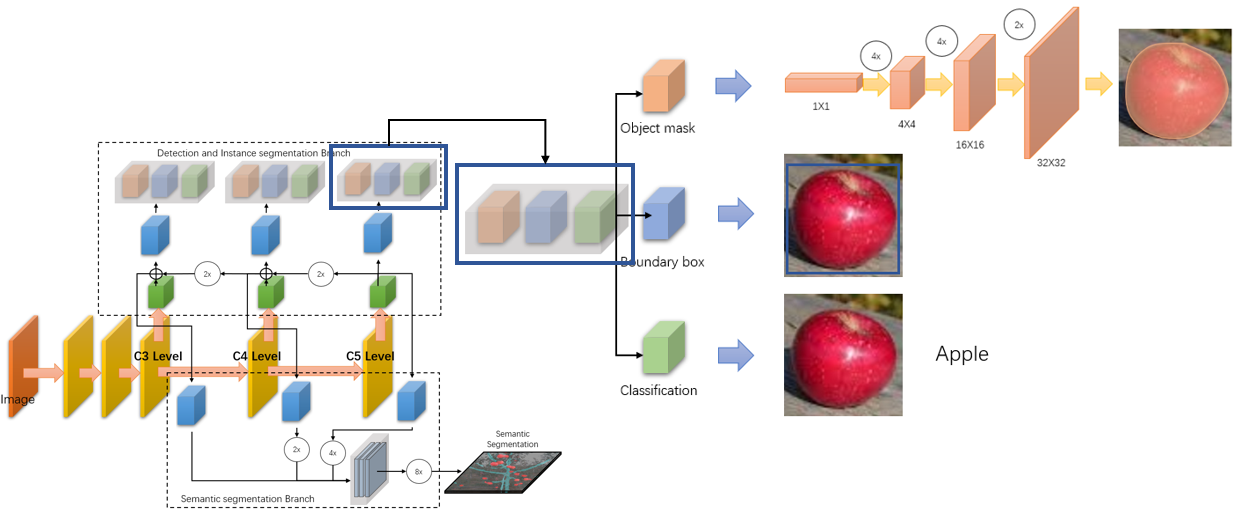}
		\caption{Network architecture of DASNet.}
		\label{fig:dasnet}
	\end{figure}
	The DASNet is utilised to perform the visual perception task in unstructured orchard environments. DASNet improves the architecture of one-stage network by combining the detection, instance segmentation, and semantic segmentation into a single network model. The detection branch of network utilises a three-level Feature Pyramid Network (FPN) to fuse the information of feature maps from deeper level to the shallower level of the network. In each level of FPN in DASNet model, an Atrous Spatial Pyramid Pooling (ASPP) is applied to encode the multi-scale features of objects into output feature maps. There are two fixed-size anchor boxes which are signed at output branch of each level of FPN (total six anchor boxes in 3-level FPN). Each output branch predicts the class, boundary box, and object mask for each of the objects. The overall design of the detection branch of DASNet is shown in Figure \ref{fig:dasnet} and more details of DASNet model can refer to the reference \cite{2019DasNet-v2}. DASNet utilises Resnet-50 as backbone since previous of our works suggest that Resnet-50 achieves the balance between performance and computation efficiency.
	
	The semantic segmentation branch is grafted at detection branch of DASNet, which receives the feature maps from C3, C4, and C5 level. To keep consistency of the feature maps from different level of the network, the C5 and C4 level of feature maps are 4$\times$ upsampled and 2$\times$ upsampled to match the size of the feature maps from the C3 level. The fusion between different level is achieved by concatenation operation. The concatenated feature tensor is then 8$\times$ upsampled to form the segmentation results. The detection branch of DASNet outputs the detection and instance segmentation results of fruits while semantic segmentation branch outputs the semantic segmentation results of the branch/trunk. The combination of the detection branch and semantic segmentation branch forms the final output of the DASNet. 
	
	\subsubsection{Post-processing}
	Due to the noisy background in unstructured orchard environments, accurate segmentation of branch/trunk is still a challenging task. Therefore, DASNet is expected to segment the major branch/trunk of trees while ignoring the other small branch. A region-connection analysis by using OpenCV API function is applied after DASNet to remove the small region the segmentation results of the branch. Except for the branch, other elements within the orchard environment such can also affect the operation of the manipulator. Thus, the objects which are identified as not belonging to the branch/trunk are also be included in the following environment modelling.

	\subsection{Visual Modelling of Environments}	
	\subsubsection{Modelling of Scenarios}
	\begin{figure}[h]
		\centering
		\includegraphics[width=\textwidth]{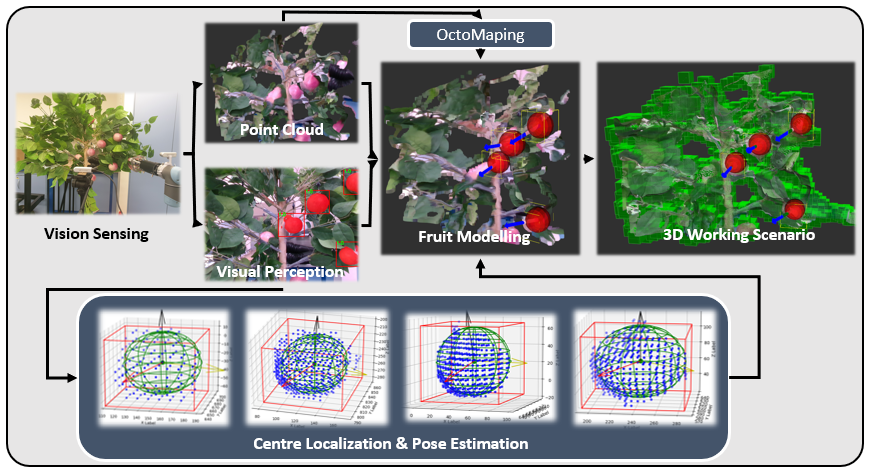}
		\caption{Workflow of working scenario modelling and fruit modelling in the laboratory with simulated fake apple tree setting.}
		\label{fig:workflow_scenario}
	\end{figure}
	From the environment perception module, the semantic labelling of fruits, branch, and other elements are returned as the form of the image binary mask. Combining the information of depth map from the RGB-D camera, the point clouds are assigned for each class of objects correspondingly. However, the 3D mapping of the environment based on point clouds is not efficient \cite{2018semanticOCTO}. Point clouds always include many noisy information and details which cannot be used in the path planning and obstacle avoiding. Meanwhile, the large number of points within the RGB-D camera output could also lead to large computation consumption.
	
	OctoMap mapping the 3D environment based on octrees, dividing the space with the number of small-sized voxels. Original OctoMap adopts probabilistic 3D mapping framework to minimise the error in the range measurement due to the moving of objects or robots. In the condition of apple harvesting robots, the current scenario collected by RGB-D camera is considered as in static condition in terms of guiding the following action of the manipulator. Therefore, the binary representation of occupied voxels within octrees is utilised. By setting the minimum size of the voxels, the resolution of the environment mapping can be adjusted based on the situation. In the process of modelling the trunk/branch, the point clouds of each class of objects will be firstly de-noised to minimise the possible measurement error during the sensing. Then, the OctoMap for each class of objects is constructed accordingly based on the given resolution. The workflow of fruit and scenario modelling algorithm is shown in Figure \ref{fig:workflow_scenario}.
	
	\subsubsection{Modelling of Fruits}
	\subsubsection*{a. Centre Localisation}
	\begin{figure}[h]
		\centering
		\includegraphics[width=\textwidth]{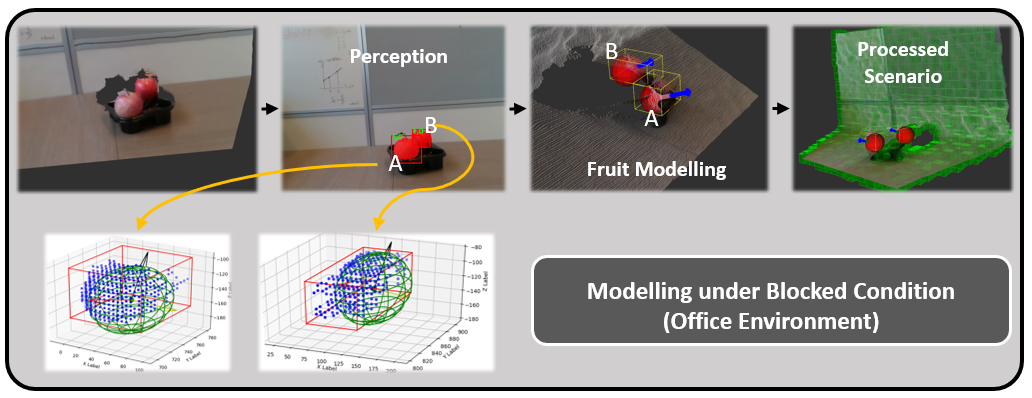}
		\caption{Working scenario modelling and modelling of partially blocked apple in the office environment.}
		\label{fig:workflow_modelling}
	\end{figure}
	A 3D-SHT algorithm is developed in this work to estimate the optimal centre position of the fruits from the obtained point clouds of each fruit mask. We assume the geometry of the apple fruit is sphere, and the equation of sphere can be expressed as:
	\begin{equation}\label{eq:shpere}
	(x-c_{x})^{2}+(y-c_{y})^{2}+(z-c_{z})^{2}=r^{2}
	\end{equation}
	$c_{x}$, $c_{y}$, $c_{z}$, and $r$ are the centre and radius of the sphere, respectively. 3D-SHT extend the Circle Hough Transform (CHT) into 3D case, applying vote framework to estimate the most possibly parameters of a sphere which can indicate the actual position and size of detected fruits within scenarios. We firstly spatially uniformed sampling the point clouds of each target to reduce the computational complexity and noise. Then, based on the distribution of point clouds, a searching range of each parameters are calculated accordingly. The searching range of each parameters will be discretized based on given resolution. For each of point within the point clouds, we calculate the corresponding value of radius $r_{est}$ for each possible pair of $c_{x}^{p}$, $c_{y}^{q}$, and $c_{z}^{n}$ within the searching range. 
	\begin{equation}\label{eq:r_est}
	r_{est}^{i}=(x_{i}-c_{x}^{p})^{2}+(y_{i}-c_{y}^{q})^{2}+(z_{i}-c_{z}^{n})^{2}
	\end{equation}
	\begin{equation}\label{eq:vote}
	if\ r_{est}^{i} \in r_{k}\ while\ r_{k} \in r_{accept},\  Grid\,[c_{x}^{p},c_{y}^{q}, c_{z}^{n}, r_{k}]+=1
	\end{equation}
	If the calculated value of radius belong to the interval of searching range $r_{accept}$ of acceptable radius, we add one vote on the corresponding parameter pair of $c_{x}^{p}$, $c_{y}^{q}$, $c_{z}^{n}$, and $r_{k}$. Finally, the parameter pair with the highest number of vote generates the estimated sphere of detected fruits.
	
	\subsubsection*{b. Pose Estimation}
	Based on estimated optimal centre of fruits and distribution of point clouds of each mask, a pose estimation algorithm is utilised to calculate the Euler-angle of each fruits within the scenario. We assume the point clouds (number of points is $n$) which is identified belong to a fruits is the visualised or unblocked partition of the fruits from the current view-angle of the RGB-D camera. According to the pose of each points within the point clouds based on the optimal centre of detected fruits, the optimal access pose of the manipulator to this fruits can be estimated. The parametric equation of a sphere is expressed as follow:
    \begin{equation}
    \label{eq:pe_sphere}
    \left\{
    \begin{aligned}
    & x=Rcos\theta sin\varphi, \\
    & y=Rsin\theta sin\varphi, \\
    & z=Rcos\varphi. \\
    \end{aligned}
    \right.
    \end{equation}
    In Eq \ref{eq:pe_sphere}, $R$ equals to $\sqrt{x^{2}+y^{2}+z^{2}}$, and $\theta$ and $\varphi$ is in the range of [0,2$\pi$]. We calculate angle $\theta_{i}$ and $\varphi_{i}$ of point $p_{i}$ which belongs to the point clouds, to estimate the unblock pose of fruits under the current view-angle of the RGB-D camera. For chosen point $p_{i}(x_{i},y_{i},z_{i})$ and centre of estimated sphere $C(c_{x},c_{y},c_{z})$ of a fruit mask, we have:
    \begin{equation}\label{eq:p_transform}
    p_{c}^{i}=(x_{c}^{i},y_{c}^{i},z_{c}^{i})=(x_{i}-c_{x},y_{i}-c_{y},z_{i}-c_{z})
    \end{equation}
    Then, the angle $\theta_{i}$ can be calculated through the function shown as follow:
    \begin{equation}\label{eq:theta_calculation}
    \theta_{i}=Atan2(\frac{y_{c}^{i}}{R_{xy}^{i}},\frac{x_{c}^{i}}{R_{xy}^{i}}), \ R_{xy}^{i}=R^{i}sin\varphi_{i}=\sqrt{(x_{c}^{i})^{2}+(y_{c}^{i})^{2}}
    \end{equation}
    Similarly, the angle $\varphi_{i}$ can be calculated through the function:
    \begin{equation}\label{eq:varphi_calculationo}
    \varphi_{i}=Atan2(\frac{z_{c}^{i}}{R_{xy}^{i}},\frac{R_{xy}^{i}}{R^{i}}), \ R^{i}=\sqrt{(x_{c}^{i})^{2}+(y_{c}^{i})^{2}+(z_{c}^{i})^{2}}
    \end{equation}
    The pose of a fruits can be modelling as the ZYX-Euler angle rotation of the coordinate around the centre of the sphere. $\theta$ and $\varphi$ are the rotation angle along the Z-axis and Y-axis, respectively. The rotation matrix therefore can be expressed as:
    \begin{equation}\label{eq:R_rotation}
    R_{pose}=R_{z}(\theta) \cdot R_{y}(\varphi) \cdot R_{x}(0)=
    {
    	\left[ \begin{array}{ccc}
    	cos\theta cos\varphi & -sin\theta & cos\theta sin\varphi\\
    	sin\theta cos\varphi & cos\theta & sin\theta sin\varphi\\
    	-sin\varphi & 0 & cos\varphi
    	\end{array} 
    	\right ]}
    \end{equation}  
    While the $\theta$ and $\varphi$ are expressed as follow:
    \begin{eqnarray}\label{eq:angle_mean}
    \theta=\frac{1}{n}\sum_{i}^{n}\theta_{i},\\
    \varphi=\frac{1}{n}\sum_{i}^{n}\varphi_{i} 
    \end{eqnarray}
    To reduce the false prediction of fruit pose due to measurement error of point clouds, we limit the range of estimated angle of $\theta$ and $\varphi$ in the range of  [$-\frac{1}{3}\pi$,$\frac{1}{3}\pi$].

	\subsubsection*{c. Pose Verification}
	Due to the complex arrangement of unstructured orchard environment, visual perception and modelling algorithm may affected by some unexpected factors. Therefore, a pose verification algorithm based on 3DVFH+ is utilised to verify the correctness of estimated pose and secure the manipulator during the operation. This method is to calculate the orientation of barrier in the given neighbourhood of a chosen fruit, to ensure that there is no barrier presented in the orientation of the estimated pose.
	
	We firstly initialise a 2-dimensional histogram $H$ to record the orientation angle $\theta$ and $\varphi$ of the barrier (branch or other elements) to the centre of fruits with a given resolution. Then we search the obstacles within neighbourhood range of $r$ (in mm) of the target fruit. That is, for a barrier $B_{i}$ within the range, the orientation angle $\theta_{i}$ and $\varphi_{i}$ to the centre $C$ of fruits are calculated by using Eqs \ref{eq:theta_calculation} and \ref{eq:varphi_calculationo}. Then a penalty value $PV_{i}$ is added to the corresponding location at $H\,[\theta_{barrier}^{i},\varphi_{barrier}^{i}]$, which can be expressed as follow:
	\begin{equation}\label{eq:vfh_histogram}
	H[\theta_{barrier}^{i},\varphi_{barrier}^{i}]+=PV_{i},\ while\ PV_{i}=\alpha*K_{i}
	\end{equation}
	The $\alpha$ and $K$ are the class term and distance term of barrier penalty which are expressed as follow:
	\begin{equation}
	\alpha=\left\{
	\begin{aligned}
	& 1,\ objects\in branch/trunk, \\
	& 0.5,\ objects\in other\,element, \\
	\end{aligned}
	\right.
	\end{equation}
	\begin{equation}
    K_{i}=\frac{1}{log_{\beta}d_{i}},\ while\ d_{i}=\lVert B_{i}-C\rVert
	\end{equation}
	$\beta$ is a constant to adjust the penalty value related to the distance penalty term. We set $\beta$ and neighbourhood range $r$ equals to 50 and 200, respectively. Then, based on the estimated pose $\theta_{pose}$ and $\varphi_{pose}$ of the chosen fruit, the sum penalty value at $H[\theta_{pose},\varphi_{pose}]$ is used to calculate the confidence rate $L$, which is expressed as:
	\begin{equation}\label{eq:confident_vari}
	L=\frac{1}{1+exp(H[\theta_{pose},\varphi_{pose}])}
	\end{equation}
	Eq \ref{eq:confident_vari} map the confidence rate of estimated pose of a target into the range of [0,1]. Therefore, the confidence rate of estimated pose of each fruits can be represented in form of the probability value. Given a threshold $\tau$ to filter the under-estimated target, which is:
	\begin{equation}\label{eq:can_pick}
	\left\{
	\begin{aligned}
	& can\_pick:\ True,\ L>=\tau \\
	& can\_pick:\ False, L<\tau\\
	\end{aligned}
	\right.
	\end{equation}
	We set $\tau$ as 0.6, and the fruits with higher confidence rate will be assigned as priority in the picking sequence. Furthermore, we should notice that Eqs \ref{eq:vfh_histogram} to \ref{eq:can_pick} only consider the geometry constraint in the environment, other constraints such as robotic working space can be further added into this framework, which can be expressed as:
	\begin{equation}
	H=\sum PV_{constraint}
	\end{equation}

    \section{Experiment and Discussion}
    \subsection{Implementation Details}
    \begin{figure}[h]
    	\centering
    	\includegraphics[width=\textwidth]{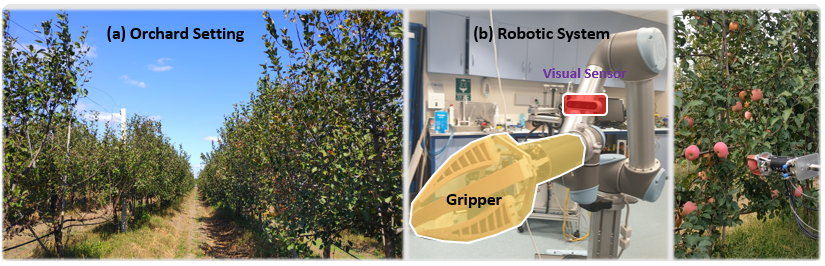}
    	\caption{(a) The orchard setting and (b) the harvesting robot.}
    	\label{fig:experiment}
    \end{figure}
     The experiment was conducted in the apple orchard, which is located at Qingdao (China), from 10th to 25th of November in the year 2019. There were 800 of RGB images which are collected from the previous in-field experiments of robotic apple harvesting. Subsets of this RGB image dataset were applied to train the model of the DASNet, while the rest RGB images were used to perform the evaluation on network performance. The model of DASNet is programmed and implemented by using Tensorflow-1.11 in python and trained on the Nvidia GTX-1080Ti. The rest programming of visual perception and modelling algorithm are implemented in ROS-kinetic on Ubuntu 16.04 and visualised by using RVIZ. The depth camera applied in the experiment is Intel-RealSense D435. The in-field implementation of the visual perception and modelling algorithm was computed on Nvidia Jetson TX2. The orchard setting and manipulator are shown in Figure \ref{fig:experiment}.
    
    \subsection{Evaluation on Environment Perception}
    \subsubsection{Evaluation Methods}
    The evaluation of environment perception module in terms of object detection, instance segmentation, and semantic segmentation are accomplished by using $F_{1}$ score, Mean Intersection of Union (MIoU), respectively. The $F_{1}$ score is used to measure the detection performance of the network by evaluating the value of $Precision$ and $Accuracy$, which are expressed as follow.
    \begin{equation}
    Precision=\frac{TruePositive(TP)}{TruePositive(TP)+FalsePositive(FP)}
    \end{equation}
    \begin{equation}
    Recall=\frac{TP}{TP+FalseNegative(FN)}
    \end{equation}
    \begin{equation}
    F_{1}=\frac{2\times\ Precision \times\ Recall}{Precision+Recall}
    \end{equation}
    The MIoU evaluates the quality of segmentation by measuring the rate between intersection and union of two subsets, which are network prediction of segmentation and ground-truth in this case.  The expression of MIoU is shown as follow \cite{2017MIoU}:
    \begin{equation}
    MIoU=\frac{1}{k}\sum_{i=1}^{k}\frac{p_{ii}}{\sum_{j=1}^{k}p_{ij}+\sum_{j=1}^{k}p_{ji}-p_{ii}}
    \end{equation}
    The $p_{ij}$ and $p_{ji}$ stand the false positive and false negative of the network prediction of segmentation, respectively.
    
    \subsubsection{Comparison with Other Methods}
    The performance of DASNet model in terms of detection, instance segmentation, and semantic segmentation are compared with YOLO (and YOLO-tiny), faster RCNN,  mask RCNN, and FCN-8s, which are shown in table as follow.
    \begin{table}[h]
    	\centering
    	\caption{Comparison of performance between different network model}
    	\begin{tabular}{ccccccc}
    		\hline
    		\textbf{Index}&\textbf{Model}&$F_{1}$ score&$MIoU_{F}$&$MIoU_{B}$\\
    		\hline
    		1&DaSNet (Resnet-50) &${0.833}$&${0.857}$&${0.792}$\\
    		\hline
    		2&YOLO-V3 (Darknet-53) &${0.811}$&${-}$&${-}$\\
    		\hline
    		3&YOLO-Tiny (Darknet-18) &${0.787}$&${-}$&${-}$\\
    		\hline
    		4&Faster RCNN (Resnet-101) &${0.834}$&${-}$&${-}$\\
    		\hline
    		5&Mask RCNN (Resnet-101) &${0.838}$&${0.871}$&${-}$\\
    		\hline
    		6&FCN-8s (Resnet-101) &${-}$&${-}$&${0.767}$\\
    		\hline
    	\end{tabular}
    	\label{table:perception_comparison}
    \end{table}
    \begin{table}[h]
    	\centering
    	\caption{Comparison of computation efficiency between network models}
    	\begin{tabular}{ccccccc}
    		\hline
    		\textbf{Device}&\textbf{Model}&\textbf{Time}\\
    		\hline
    		Jetson-TX2 &DaSNet-V2 (Resnet-50) &${0.31s}$\\
    		\hline
    		Jetson-TX2 &YOLO (Darknet-53) &${0.27s}$\\
    		\hline
    		Jetson-TX2 &Faster RCNN (Resnet-101) &${1.2s}$\\
    		\hline
    		Jetson-TX2 &Mask RCNN (Resnet-101) &${1.2s}$\\
    		\hline
    	\end{tabular}
    	\label{table:network_efficiency}
    \end{table}
    
    \begin{figure}
    	\centering
    	\includegraphics[width=\textwidth]{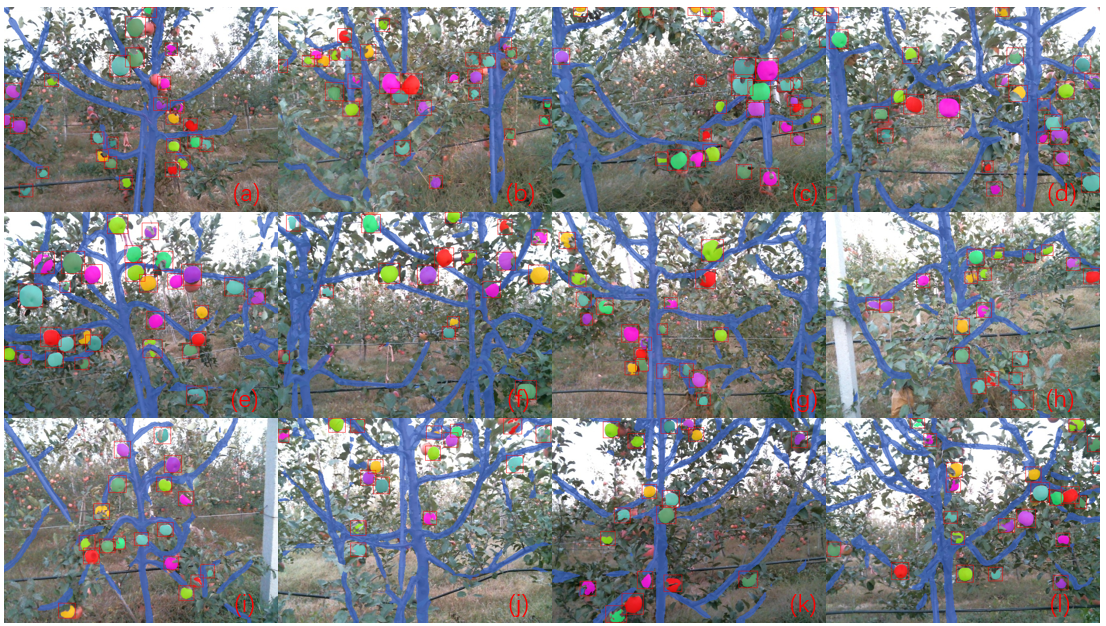}
    	\caption{Visual perception results in apple orchard by using DASNet.}
    	\label{fig:perception_results}
    \end{figure}
    The $MIoU_{F}$ and $MIoU_{B}$ in Table \ref{table:perception_comparison} stand the accuracy of instance segmentation on fruits and semantic segmentation on branch, respectively. From the experimental results, DASNet improves the detection performance of one-stage network compared to the YOLO from 0.811 to the 0.833, and achieves comparable performance compared to the two-stage detection network faster RCNN and mask RCNN, which are 0.833, 0.834 and 0.838, respectively. In terms of the instance segmentation of detected objects, DASNet also shows a similar accuracy compared to the mask RCNN, which are 0.857 and 0.871, respectively. The potential reason for the higher accuracy of instance segmentation on mask RCNN is that the two-stage detection network can apply RPN and ROI re-alignment to segment the corresponding area of the objects accurately. On another hand, although DASNet applies ASPP to encode multi-scale features, multi-scale features may also introduce noise and lead to a lower accuracy of segmentation. From the following experimental results, DASNet is capable of providing accurate segmentation of fruits. In terms of the semantic segmentation, DASNet achieves similar performance compared to the FCN-8s model, which are 0.802 and 0.767, respectively. However, due to the complex conditions in the implementation of robotic harvesting in the unstructured orchard, to accurately segment the branch/trunk from the noisy background is still a challenging task. Therefore, environment modelling algorithm considers both inputs from the branch/trunk and elements of other objects within the working scenarios. The working scenarios which are processed by the visual perception algorithm are shown in Figure \ref{fig:perception_results}. In addition, the comparison of the computation efficiency of different network models is given in Table \ref{table:network_efficiency}.
    
    \subsection{Evaluation on Fruits Modelling}
    The evaluation of fruits modelling algorithm is achieved by measure its accuracy and robustness along the distance within the working range of the harvesting robot. The accuracy evaluation of modelling algorithm is conducted by manual marking the correct-estimated objects and wrong-estimated objects in each scenario. The robustness evaluation measures the fluctuating range (Standard Deviation (SD)) of estimated geometries property and object poses in multiple running (repeat 5-10 times in each scenario) of the same scenario. Both tests are evaluated along the distance from the depth camera to the objects.
        
    \subsubsection{Localisation and Pose Estimation}
    The operating range of the vision system is from 0.3m to 0.7m along the X-axis of the robot coordinate (the minimum steady operation distance of Intel RealSense D435 is 0.2m). To evaluate the performance of the developed system, we extend the maximum test range of visual system up to 1.2m (in the range of >0.9m). Table \ref{table:fruit_modelling_centre} shows the experimental result on 3D-SHT algorithm. Table \ref{table:fruit_modelling_PM} shows the Average Size of Boundary Box (ASoBBx), Average Number of Pixels (ANoP), and Average Number of Computation Candidate (ANoCC) of each object within images along the distance.
    
    \begin{table}[h]
    	\centering
    	\caption{Experimental results on geometry estimation of Fruit Modelling}
    	\begin{tabular}{ccccccc}
    		\hline
    		\textbf{Distance (m)} &0.3-0.5&0.5-0.7&0.7-0.9&>0.9\\
    		\hline
    		\textbf{Accuracy} &0.955&0.925&0.85&0.775\\
    		\hline
    		\textbf{SD(centre) (mm)} &3.2&5.5&7.6&13.6\\
    		\hline
    		\textbf{SD(radius) (mm)} &3.8&6.2&11.6&20.7\\
    		\hline
    	\end{tabular}
    	\label{table:fruit_modelling_centre}
    \end{table}

    \begin{table}[h]
    	\centering
    	\caption{Experimental results on pose estimation of Fruit Modelling}
    	\begin{tabular}{ccccccc}
    		\hline
    		\textbf{Distance (m)} &0.3-0.5&0.5-0.7&0.7-0.9&>0.9\\
    		\hline
    		\textbf{Accuracy} &0.923&0.885&0.775&0.725\\
    		\hline
    		\textbf{SD($\theta$) ($^{\circ}$)} &5.2&8.6&13.7&22.5\\
    		\hline
    		\textbf{SD($\varphi$) ($^{\circ}$)} &4.9&7.6&12.5&23.2\\
    		\hline
    	\end{tabular}
    	\label{table:fruit_modelling_pose}
    \end{table}

    \begin{table}[h]
    	\centering
    	\caption{Average size of boundary box and number of pixels of objects within images along the distance}
    	\begin{tabular}{ccccccc}
    		\hline
    		\textbf{Distance (m)} &0.3-0.5&0.5-0.7&0.7-0.9&>0.9\\
    		\hline
    		\textbf{ASoBBx} &>80&$\approx$80&$\approx$50&<50\\
    		\hline
    		\textbf{ANoP} &>2$\times10^{3}$&$\approx$2$\times 10^{3}$&$\approx$800&<800\\
    		\hline
    		\textbf{ANoCC} &>200&$\approx$200&$\approx$80&<80\\
    		\hline
    	\end{tabular}
    	\label{table:fruit_modelling_PM}
    \end{table}
    
    Rather than utilising all point candidates of an object to compute the geometry properties which is computation inefficient and time-consuming, a voxel downsampling algorithm is utilised. This step generates the computation candidates of each object, which lead to more efficient computation. Therefore, ANoCC records the average number of points which are used to calculate for each of the objects in a different range of operating distance.
    
    From the experiment results shown in Table \ref{table:fruit_modelling_centre}, it can be seen that the accuracy of the 3D-SHT algorithm shows a decrease with increasing of the distance from the depth camera to objects. Also, the fluctuating range of estimated parameters of objects becomes larger with the increase of distance. The reason that leads to the results is due to the changing of object scale in different distance within images. As shown in Table \ref{table:fruit_modelling_PM}, with the increase of distance from depth camera to the objects, the ASoBBX, ANoP, and ANoCC show a dramatic decrease. The limited number of point candidates lack the capability of describing the objects with enough information and details, which may lead to false prediction. From the experiment results (Table \ref{table:fruit_modelling_centre}) within the distance of 0.7m, centre estimation algorithm can robustly perform the task with acceptable fluctuating of estimation (around 5.5mm). The accuracy of centre estimation algorithm in the range of 0.3-0.5m and 0.5-0.7m are 0.955 and 0.925, respectively.
    
    \begin{figure}[h]
    	\centering
    	\includegraphics[width=\textwidth]{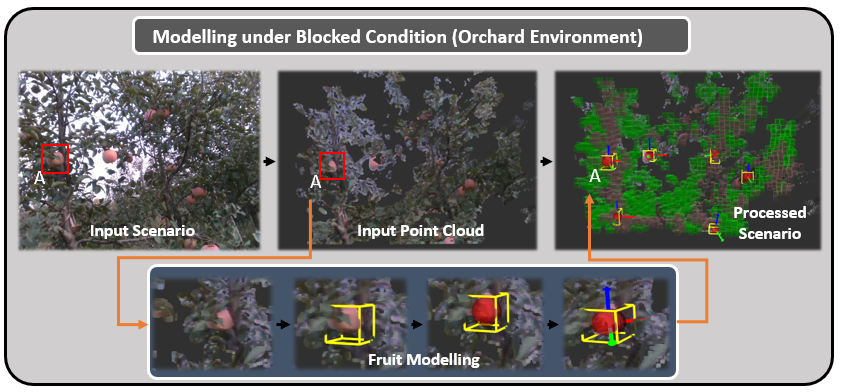}
    	\caption{Working scenario modelling and modelling of partially blocked apple in the unstructured orchards.}
    	\label{fig:block_orchard}
    \end{figure}
    The experimental results of the fruit pose estimation algorithm of the objects are shown in Table \ref{table:fruit_modelling_pose}. Similar to the case of centre estimation of fruits pose estimation algorithm can work robustly under the range of 0.7m. The accuracy and SD of $\theta$ and $\varphi$ between 0.3-0.5m and 0.5-0.7m are 0.923, 5.2$^{\circ}$, 4.9$^{\circ}$ and 0.885, 8.6$^{\circ}$ and 7.6$^{\circ}$, respectively. With the increase of distance from the depth camera (decrease of ANoCC), the accuracy and robustness of pose estimation show a significant decrease. Therefore, a 3DVFH+ based pose verification algorithm is utilised to check the correctness of the estimated pose, which can ensure the safety of manipulator.
    
    \subsubsection{Modelling of partially Blocked Objects}
    \begin{figure}[h]
    	\centering
    	\includegraphics[width=\textwidth]{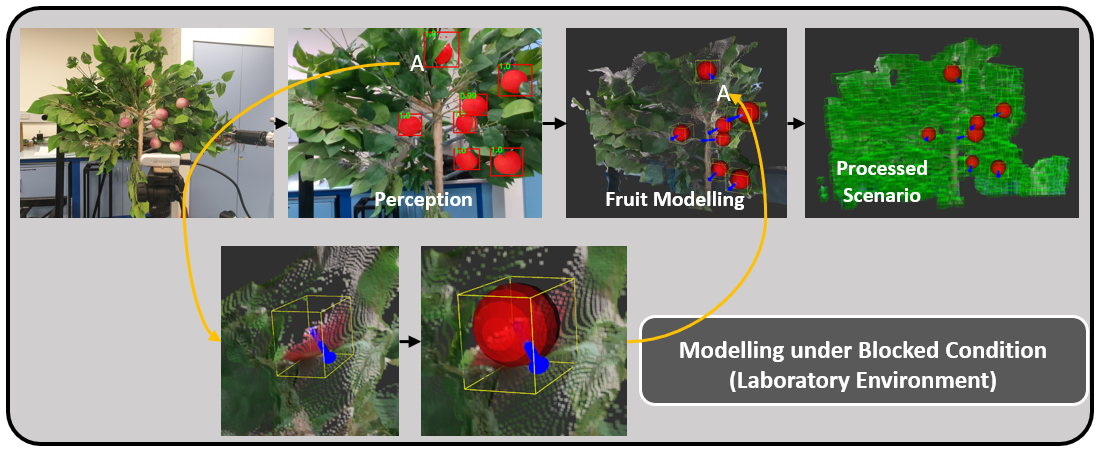}
    	\caption{Working scenario modelling and modelling of partially blocked apple in the laboratory environment.}
    	\label{fig:block_lab}
    \end{figure}
    Modelling of fruits which are partially blocked by other objects is an important task in the robotic harvesting. With the advance of deep-learning based instance segmentation method, DASNet can robustly perform instance segmentation on such partially Blocked Objects. This experiment evaluates the centre and pose estimation algorithm on fruit in partially blocked conditions. As shown in Figures \ref{fig:workflow_modelling}, \ref{fig:block_orchard}, and \ref{fig:block_lab}, the fruit modelling algorithm can efficiently perform the centre localisation and pose estimation when partially surface of fruits are blocked by other elements within the environment. To avoid the failure of modelling algorithm when severely blocked fruits are presented, the fruits without the sufficient number of point candidate will be removed from modelling list. In addition, other factors such as sensor measurement noise could also heavily affect the results of vision processing. The performance of the developed vision algorithm and correspond offset solutions to deal with the variance within orchard are included in the following section.  
        
    \subsection{Evaluation on Overall System}
    \subsubsection*{a. Point Clouds Processing}
    \begin{figure}[h]
    	\centering
    	\includegraphics[width=\textwidth]{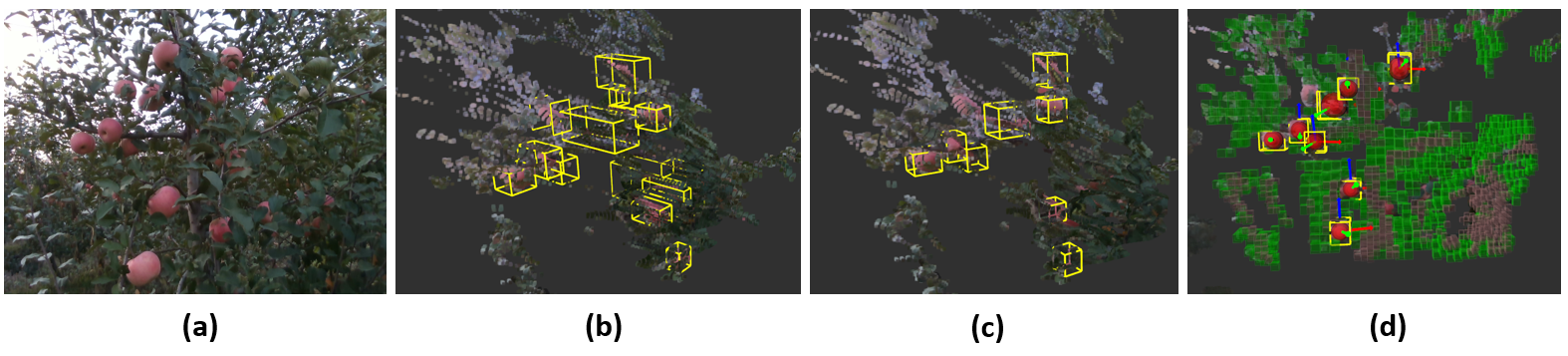}
    	\caption{(a) the input image, (b) the original object boundary boxes of fruits, (c) the object boundary boxes of fruits after filtering, and (d) the modelling results of the scenarios.}
    	\label{fig:pcp}
    \end{figure}
    In the unstructured orchard environment, the depth camera can be affected by the complex arrangement of elements within the scenarios, such as inaccurate depth sensing of points due to the mismatch between the RGB image and the depth image. In addition, the depth-sensing of the objects can be affected by the adjacent objects. Such defect could severely affect the working of the centre localisation and pose estimation algorithm (as shown in Figure \ref{fig:pcp}). Therefore, the point cloud of objects will firstly be de-noised before fruit modelling. That is, the points of an object will be classified as inlier or outlier based on Euclidean distance. Then the points which are classified as outliers will be deleted from the point list. Moreover, based on the size of the object boundary boxes, the objects without the sufficient number of points or with severely in-balance length in different axis (X, Y, Z) will be deleted from object list (as shown in (b) and (c) in Figure \ref{fig:pcp}). In the recent work of robotic harvesting of strawberry \cite{2019DLW_4}, similar issues of the depth sensing are reported. Their work applied a Density-Based Spatial Clustering (DBSC) to process the point clouds. Our implementation of point cloud denoising is more concise and computation efficient. Meanwhile, experiment results also indicate that our method is efficient in the case of robotic harvesting in unstructured orchards.
    
    \subsubsection*{b. Experiment Results in Orchard} 
    \begin{figure}[h]
    	\centering
    	\includegraphics[width=\textwidth]{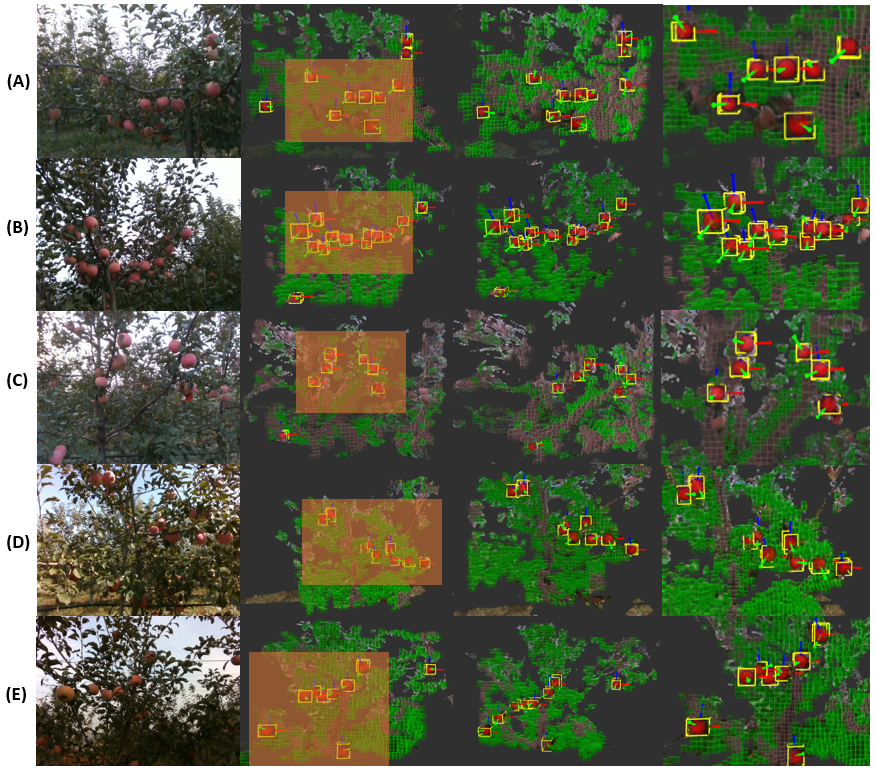}
    	\caption{The visualised working scenarios of the unstructured orchard environments by using visual perception and modelling algorithm (gray boxes stand the branch/trunk class).}
    	\label{fig:results}
    \end{figure}
    The experiments are conducted in the apple orchard located at Qingdao, China. The plant setting of the orchard is shown in Figure \ref{fig:experiment}. The time period of the experiment is in the range of 10:00 to 16:00 of days in the apple harvest season. As shown the experiment results of processing scenarios within apple orchard in Figure \ref{fig:results}, the developed visual perception and modelling algorithm can efficiently sensing the fruit and branch from the RGB image and modelling the 3D working scenario from the depth map.
    
    \section{Conclusion}
    In this work, A visual perception and modelling algorithm for robotic harvesting of apple in the unstructured orchard is developed. The visual perception comprises a multi-purpose network DASNet to perform detection, instance segmentation of fruits and semantic segmentation of working scenarios. The visual modelling algorithm performs the centre localisation and pose estimation of fruits and modelling the elements within the scenarios to guide the motion of the manipulator. The experiment results showed that visual perception and modelling algorithm could accurately detect and localise the fruits, and modelling working scenarios in real orchard environments. The $F_{1}$ score and MIoU of DASNet on fruit detection and segmentation are 0.833 and 0.852, respectively. The accuracy of centre localisation and pose estimation are 0.955 and 0.923, respectively. Furthermore, 3DVFH+ algorithm based pose verification algorithm is developed to ensure the safety of manipulator in operation.

	\section*{Acknowledgement}
	This work is supported by ARC ITRH IH150100006 and THOR TECH PTY LTD. We acknowledge Zhuo Chen for her assistance in preparation of this work. 
	
	\bibliographystyle{ieeetr}
	\bibliography{Reference}

\begin{thebibliography}{10}

\bibitem{2016TMW_review2}
Y.~Zhao, L.~Gong, Y.~Huang, and C.~Liu, ``A review of key techniques of
  vision-based control for harvesting robot,'' {\em Computers and Electronics
  in Agriculture}, vol.~127, pp.~311--323, 2016.

\bibitem{2016colorgeof}
T.~T. Nguyen, K.~Vandevoorde, N.~Wouters, E.~Kayacan, J.~G. De~Baerdemaeker,
  and W.~Saeys, ``Detection of red and bicoloured apples on tree with an rgb-d
  camera,'' {\em Biosystems Engineering}, vol.~146, pp.~33--44, 2016.

\bibitem{lin2019colorgeof}
G.~Lin, Y.~Tang, X.~Zou, J.~Xiong, and Y.~Fang, ``Color-, depth-, and
  shape-based 3d fruit detection,'' {\em Precision Agriculture}, pp.~1--17,
  2019.

\bibitem{wang2018unsupervised}
Y.~Wang and L.~Xu, ``Unsupervised segmentation of greenhouse plant images based
  on modified latent dirichlet allocation,'' {\em PeerJ}, vol.~6, p.~e5036,
  2018.

\bibitem{2012TMW_review1}
K.~Kapach, E.~Barnea, R.~Mairon, Y.~Edan, and O.~Ben-Shahar, ``Computer vision
  for fruit harvesting robots--state of the art and challenges ahead,'' {\em
  International Journal of Computational Vision and Robotics}, vol.~3, no.~1/2,
  pp.~4--34, 2012.

\bibitem{2012TMW_review3}
A.~Vibhute and S.~Bodhe, ``Applications of image processing in agriculture: a
  survey,'' {\em International Journal of Computer Applications}, vol.~52,
  no.~2, 2012.

\bibitem{2015faster-rcnn}
S.~Ren, K.~He, R.~Girshick, and J.~Sun, ``Faster r-cnn: Towards real-time
  object detection with region proposal networks,'' in {\em Advances in neural
  information processing systems}, pp.~91--99, 2015.

\bibitem{2018yolov3}
J.~Redmon and A.~Farhadi, ``Yolov3: An incremental improvement,'' {\em arXiv
  preprint arXiv:1804.02767}, 2018.

\bibitem{2015FCN}
J.~Long, E.~Shelhamer, and T.~Darrell, ``Fully convolutional networks for
  semantic segmentation,'' in {\em Proceedings of the IEEE conference on
  computer vision and pattern recognition}, pp.~3431--3440, 2015.

\bibitem{2017mask-rcnn}
K.~He, G.~Gkioxari, P.~Doll{\'a}r, and R.~Girshick, ``Mask r-cnn,'' in {\em
  Proceedings of the IEEE international conference on computer vision},
  pp.~2961--2969, 2017.

\bibitem{2016DLW_1}
I.~Sa, Z.~Ge, F.~Dayoub, B.~Upcroft, T.~Perez, and C.~McCool, ``Deepfruits: A
  fruit detection system using deep neural networks,'' {\em Sensors}, vol.~16,
  no.~8, p.~1222, 2016.

\bibitem{2017DLW_2}
S.~Bargoti and J.~Underwood, ``Deep fruit detection in orchards,'' in {\em 2017
  IEEE International Conference on Robotics and Automation (ICRA)},
  pp.~3626--3633, IEEE, 2017.

\bibitem{2019DLW_3}
Y.~Yu, K.~Zhang, L.~Yang, and D.~Zhang, ``Fruit detection for strawberry
  harvesting robot in non-structural environment based on mask-rcnn,'' {\em
  Computers and Electronics in Agriculture}, vol.~163, p.~104846, 2019.

\bibitem{2019DLW_4}
Y.~Ge, Y.~Xiong, G.~L. Tenorio, and P.~J. From, ``Fruit localization and
  environment perception for strawberry harvesting robots,'' {\em IEEE Access},
  vol.~7, pp.~147642--147652, 2019.

\bibitem{2019DLW_5}
Y.~Tian, G.~Yang, Z.~Wang, H.~Wang, E.~Li, and Z.~Liang, ``Apple detection
  during different growth stages in orchards using the improved yolo-v3
  model,'' {\em Computers and electronics in agriculture}, vol.~157,
  pp.~417--426, 2019.

\bibitem{2019DLW_6}
A.~Koirala, K.~Walsh, Z.~Wang, and C.~McCarthy, ``Deep learning for real-time
  fruit detection and orchard fruit load estimation: Benchmarking of
  ‘mangoyolo’,'' {\em Precision Agriculture}, pp.~1--29, 2019.

\bibitem{2019DLW_7}
G.~Lin, Y.~Tang, X.~Zou, J.~Xiong, and J.~Li, ``Guava detection and pose
  estimation using a low-cost rgb-d sensor in the field,'' {\em Sensors},
  vol.~19, no.~2, p.~428, 2019.

\bibitem{2017DLW_8}
Y.~Li, Z.~Cao, Y.~Xiao, and A.~B. Cremers, ``Deepcotton: in-field cotton
  segmentation using deep fully convolutional network,'' {\em Journal of
  Electronic Imaging}, vol.~26, no.~5, p.~053028, 2017.

\bibitem{2019LedNet}
H.~Kang and C.~Chen, ``Fast implementation of real-time fruit detection in
  apple orchards using deep learning,'' {\em Computers and Electronics in
  Agriculture}, p.~105108, 2019.

\bibitem{2019DasNet-v1}
H.~Kang and C.~Chen, ``Fruit detection and segmentation for apple harvesting
  using visual sensor in orchards,'' {\em Sensors}, vol.~19, no.~20, p.~4599,
  2019.

\bibitem{2019DasNet-v2}
H.~Kang and C.~Chen, ``Fruit detection, segmentation and 3d visualisation of
  environments in apple orchards,'' {\em arXiv preprint arXiv:1911.12889},
  2019.

\bibitem{2019SPRNet}
J.~Yao, Z.~Yu, J.~Yu, and D.~Tao, ``Single pixel reconstruction for one-stage
  instance segmentation,'' {\em arXiv preprint arXiv:1904.07426}, 2019.

\bibitem{2018DLW_review}
A.~Kamilaris and F.~X. Prenafeta-Bold{\'u}, ``Deep learning in agriculture: A
  survey,'' {\em Computers and electronics in agriculture}, vol.~147,
  pp.~70--90, 2018.

\bibitem{1989mapping_1}
Y.~Roth-Tabak and R.~Jain, ``Building an environment model using depth
  information,'' {\em Computer}, vol.~22, no.~6, pp.~85--90, 1989.

\bibitem{2006mapping_2}
D.~M. Cole and P.~M. Newman, ``Using laser range data for 3d slam in outdoor
  environments,'' in {\em Proceedings 2006 IEEE International Conference on
  Robotics and Automation, 2006. ICRA 2006.}, pp.~1556--1563, IEEE, 2006.

\bibitem{2009mapping_3}
R.~Hadsell, J.~A. Bagnell, D.~F. Huber, and M.~Hebert, ``Accurate rough terrain
  estimation with space-carving kernels.,'' in {\em Robotics: Science and
  Systems}, vol.~2009, 2009.

\bibitem{2010mapping_4}
B.~Douillard, J.~Underwood, N.~Melkumyan, S.~Singh, S.~Vasudevan, C.~Brunner,
  and A.~Quadros, ``Hybrid elevation maps: 3d surface models for
  segmentation,'' in {\em 2010 IEEE/RSJ International Conference on Intelligent
  Robots and Systems}, pp.~1532--1538, IEEE, 2010.

\bibitem{2007mapping_5}
M.~Magnusson, A.~Lilienthal, and T.~Duckett, ``Scan registration for autonomous
  mining vehicles using 3d-ndt,'' {\em Journal of Field Robotics}, vol.~24,
  no.~10, pp.~803--827, 2007.

\bibitem{2007mapping_6}
M.~Habbecke and L.~Kobbelt, ``A surface-growing approach to multi-view stereo
  reconstruction,'' in {\em 2007 IEEE Conference on Computer Vision and Pattern
  Recognition}, pp.~1--8, IEEE, 2007.

\bibitem{2013octomap}
A.~Hornung, K.~M. Wurm, M.~Bennewitz, C.~Stachniss, and W.~Burgard, ``Octomap:
  An efficient probabilistic 3d mapping framework based on octrees,'' {\em
  Autonomous robots}, vol.~34, no.~3, pp.~189--206, 2013.

\bibitem{2018octomap_flight}
M.~Beul, D.~Droeschel, M.~Nieuwenhuisen, J.~Quenzel, S.~Houben, and S.~Behnke,
  ``Fast autonomous flight in warehouses for inventory applications,'' {\em
  IEEE Robotics and Automation Letters}, vol.~3, no.~4, pp.~3121--3128, 2018.

\bibitem{2018semanticOCTO}
L.~Zhang, L.~Wei, P.~Shen, W.~Wei, G.~Zhu, and J.~Song, ``Semantic slam based
  on object detection and improved octomap,'' {\em IEEE Access}, vol.~6,
  pp.~75545--75559, 2018.

\bibitem{2011orchard_modelling_1}
B.~Adhikari and M.~Karkee, ``3d reconstruction of apple trees for mechanical
  pruning,'' in {\em 2011 Louisville, Kentucky, August 7-10, 2011}, p.~1,
  American Society of Agricultural and Biological Engineers, 2011.

\bibitem{2013orchard_modelling_2}
Q.~Wang and Q.~Zhang, ``Three-dimensional reconstruction of a dormant tree
  using rgb-d cameras,'' in {\em 2013 Kansas City, Missouri, July 21-July 24,
  2013}, p.~1, American Society of Agricultural and Biological Engineers, 2013.

\bibitem{2016orchard_modelling_3}
S.~Amatya, M.~Karkee, A.~Gongal, Q.~Zhang, and M.~D. Whiting, ``Detection of
  cherry tree branches with full foliage in planar architecture for automated
  sweet-cherry harvesting,'' {\em Biosystems engineering}, vol.~146, pp.~3--15,
  2016.

\bibitem{2014_3dvfh+}
S.~Vanneste, B.~Bellekens, and M.~Weyn, ``3dvfh+: Real-time three-dimensional
  obstacle avoidance using an octomap,'' in {\em MORSE 2014 Model-Driven Robot
  Software Engineering: proceedings of the 1st International Workshop on
  Model-Driven Robot Software Engineering co-located with International
  Conference on Software Technologies: Applications and Foundations (STAF
  2014), York, UK, July 21, 2014/Assmann, Uwe [edit.]}, no.~1319, pp.~91--102,
  2014.

\bibitem{2017MIoU}
A.~Garcia-Garcia, S.~Orts-Escolano, S.~Oprea, V.~Villena-Martinez, and
  J.~Garcia-Rodriguez, ``A review on deep learning techniques applied to
  semantic segmentation,'' {\em arXiv preprint arXiv:1704.06857}, 2017.

\end{thebibliography}
	\renewcommand\refname{References} 
	
\end{document}